\documentclass[letterpaper, 10 pt, conference]{ieeeconf}  % Comment this line out if you need a4paper

\IEEEoverridecommandlockouts                              % This command is only needed if 
\usepackage{graphicx}
\usepackage{cite}
\usepackage{amsmath,amssymb,amsfonts}
\usepackage{algorithmic}
\usepackage{graphicx}
\usepackage{textcomp}
\usepackage{xcolor}
\usepackage[dvipsnames]{xcolor}
\usepackage[ruled,vlined,linesnumbered]{algorithm2e}
\usepackage{bm}
\usepackage{subfig}
\usepackage{url}
\usepackage{hyperref}
\usepackage{booktabs}

\usepackage{comment}

\title{\LARGE \bf 
KRAST: Knowledge-Augmented Robotic Action Recognition with Structured Text for Vision-Language Models  
}
\author{
Son~Hai~Nguyen$^{1}$\thanks{
$^{1}$ \texttt{son-hai.nguyen@etu.univ-cotedazur.fr}},
Hyewon~Seo$^{2}$\thanks{
$^{2}$ \texttt{seo@unistra.fr}},
Diwei~Wang$^{3}$\thanks{
$^{3}$ \texttt{d.wang@unistra.fr}},
Jinhyeok~Jang$^{4}$\thanks{
$^{4}$ \texttt{jjh6297@etri.kr}}
}
\begin{document}

\maketitle
\thispagestyle{empty}
\pagestyle{empty}

%%%%%%%%%%%%%%%%%%%%%%%%%%%%%%%%%%%%%%%%%%%%%%%%%%%%%%%%%%%%%%%%%%%%%%%%%%%%%%%%
\begin{abstract}

Accurate vision-based action recognition is crucial for developing autonomous robots that can operate safely and reliably in complex, real-world environments. In this work, we advance video-based recognition of indoor daily actions for robotic perception by leveraging vision-language models (VLMs) enriched with domain-specific knowledge. We adapt a prompt-learning framework in which class-level textual descriptions of each action are embedded as learnable prompts into a frozen pre-trained VLM backbone. Several strategies for structuring and encoding these textual descriptions are designed and evaluated. Experiments on the ETRI-Activity3D dataset demonstrate that our method, using only RGB video inputs at test time, achieves over 95\% accuracy and outperforms state-of-the-art approaches. These results highlight the effectiveness of knowledge-augmented prompts in enabling robust action recognition with minimal supervision.

\end{abstract}

%%%%%%%%%%%%%%%%%%%%%%%%%%%%%%%%%%%%%%%%%%%%%%%%%%%%%%%%%%%%%%%%%%%%%%%%%%%%%%%%

\section{INTRODUCTION}

While understanding human activities from video is a fundamental requirement for intelligent systems operating in human-centered environments, robust vision-based recognition still remains a challenge. Domestic settings such as smart homes and assistive robots  highlight this need, as accurate, real-time  human activity recognition (HAR) is essential for context-aware automation and personalized elderly care. Yet, real-world environments filled with furniture and other objects are inherently occluded, cluttered, and unpredictable, which continues to impede the development of reliable solutions.

Traditional HAR approaches have relied on hand-crafted features or depth-based skeleton tracking, which often suffer from poor generalization in diverse, unconstrained in-home settings. Recent advances in deep learning, particularly convolutional and transformer-based architectures, have significantly improved the capacity to model spatiotemporal dynamics of motion from raw RGB data. However, most existing deep learning–based HAR models rely solely on visual input, making them susceptible to performance degradation in real-world scenarios involving occlusions and subtle activity variations-—such as distinguishing between sitting and lying, or reaching and pointing.

%%% Newly added
The recent success of multimodal vision–language models (VLMs) has shown a strong potential for transferring knowledge across domains by jointly leveraging visual inputs and textual descriptions. 
Building on the pre-trained reasoning capabilities of VLMs, we elaborate a lightweight fine-tuning framework that employs prompting strategies specifically tailored for vision-based human activity recognition in domestic environments. 
Specifically, we introduce prompting strategies where textual action descriptions convey structural relationships among action classes, such as hierarchical categories or taxonomies. These are then used to condition the learnable prompts to better guide the vision–language model for robust action recognition.
%%% Moved from related work
%In this work, we propose to rely on a multimodel framework that exploits  integrates visual inputs with textual descriptions via a vision-language model (VLM) tailored for vision-based human activity recognition in domestic environments, addressing these challenges through a multimodal approach that integrates visual inputs with textual descriptions via a vision-language model (VLM).
%% Previous intro
%In this work, we propose a deep learning framework tailored for vision-based human activity recognition in domestic environments, addressing these challenges through a multimodal approach that integrates visual inputs with textual descriptions via a vision-language model (VLM).
%We evaluate these strategies and demonstrate that their combination yields the best performance.
We demonstrate how this simple yet effective approach aids in  distinguishing visually similar actions commonly encountered in indoor environments.

%To address challenges arising from cluttered and subtle action variations, 
%We evaluate our approach on a publicly available dataset~\cite{FSA-CNN}, which demonstrate state-of-the-art performance in comparison with existing methods. 
%This work contributes toward building more perceptive and responsive in-home AI systems, with implications for smart robotics, ambient assisted living, and human-aware automation.

\section{Related work}
A key direction in the field involves bridging human activity recognition (HAR) with practical robotic applications by developing models and datasets that account for real-world constraints.
Toupas et al. ~\cite{Toupas2023} proposed an edge-computing pipeline that enables service robots to detect and classify common household actions, such as sitting, reaching, or walking, using onboard RGB sensors in real-time. In a similar context, the ETRI-Activity3D dataset ~\cite{FSA-CNN} provides a large-scale benchmark of elderly daily activities captured from a robot’s viewpoint, facilitating the development of vision models grounded in real-world scenarios. 

Broadly applicable vision-based action recognition models have also been extensively developed on large-scale unconstrained datasets, such as Kinetics~\cite{kinetics_2017}, UCF101~\cite{ucf101_2012}, and HMDB51~\cite{hmdb_2011}. Deep architectures such as the 3D convolutional network I3D~\cite{I3D}, the dual-pathway model SlowFast~\cite{SlowFast}, and the Transformer-based approach TimeSformer~\cite{TimeSformer} have achieved state-of-the-art performance by effectively modeling rich spatiotemporal dynamics.

More recently, multimodal extensions such as ActionCLIP~\cite{actionclip} have demonstrated the effectiveness of incorporating language supervision to enhance recognition accuracy.
%While covering a wide range of actions, including sports and outdoor actions, these models are not specifically optimized for daily indoor scenarios.
However, these models have been developed for, and trained on, general action recognition settings where the range of motions is broad and distinctions between actions are relatively clear. In contrast, indoor daily action recognition involves motions that are often subtle and highly context-dependent -- washing hands versus washing dishes, or brushing teeth versus putting on lipstick, for instance. As a result, general-purpose action recognition models often require fine-tuning and task-specific adaptation to achieve robust performance in such specialized environments.

Prompt tuning has emerged as a common strategy for adapting vision–language models (VLMs) to downstream tasks, which is also adopted in this work. Despite the development of advanced variants such as knowledge-augmented prompt tuning (KAPT)~\cite{kapt}, existing studies largely rely on empirical prompt design, and it remains unclear how the structure and semantics of the textual knowledge influence model performance. 
Current findings primarily highlight the sensitivity of VLMs to prompt formulation~\cite{actionclip,coop}, but offer limited insight into how knowledge can systematically guide prompting for improved recognition. We address this gap by designing and evaluating different strategies for conditioning textual prompts, using descriptions that capture  structural relationships among action classes.
%To address this issue, we introduce a prompting strategy where textual action descriptions convey structural relationships among action classes, such as hierarchical categories or taxonomies. These are then used to condition the learnable prompts to better guide the vision–language model for robust action recognition.

\section{Dataset and preprocessing}

\subsection{Dataset}

\begin{figure}[h]
  \centering
  \includegraphics[width=\linewidth]{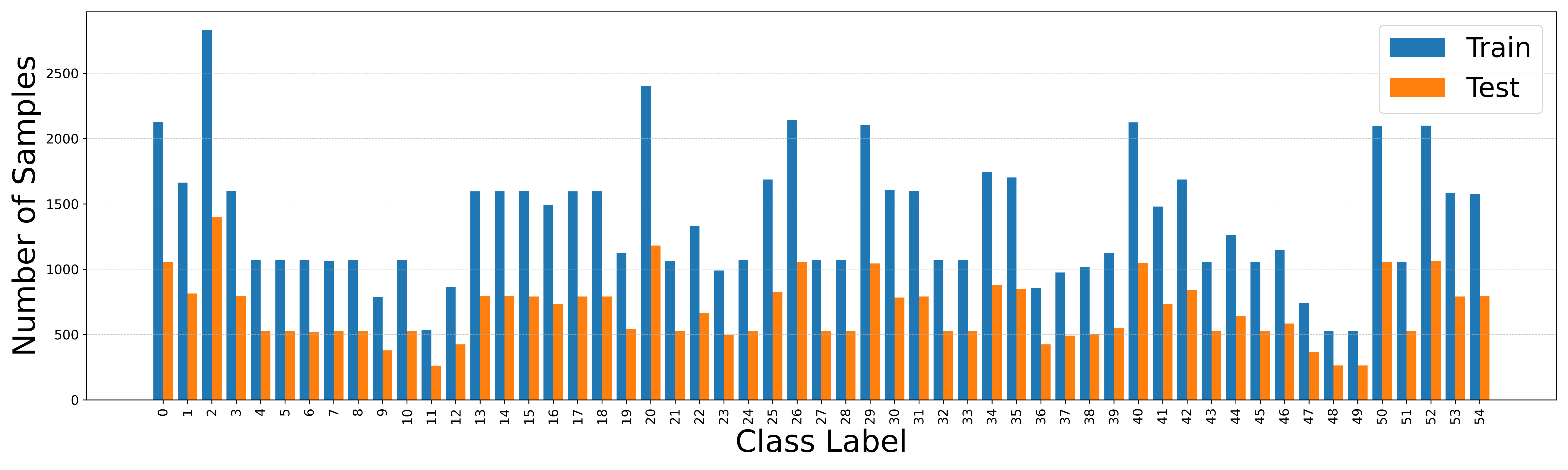}
  \caption{Statistics of per-class sample counts  for training and validation sets.}
  \label{fig:statistics}
\end{figure}

The ETRI-Activity3D dataset~\cite{FSA-CNN} is a large-scale benchmark designed for video-based action recognition, particularly in the context of elderly care. In this study, only the RGB video modality is used, recorded at a high resolution of $1920 \times 1080$ pixels with Kinect v2 sensors. The dataset includes 55 action classes: 52 daily activities and 3 human-robot interactions (e.g., waving, pointing). Fig.\ref{fig:statistics} illustrates the number of samples for each class in this dataset.
It includes 100 subjects—50 elderly adults (aged 64–88) and 50 young adults (around age 23)—with diverse video recordings from multiple angles, heights, and distances. Several actions are performed in varying contexts to increase intra-class diversity, making the dataset suitable for evaluating models in complex, real-world scenarios.

\subsection{Preprocessing}

In the preprocessing stage, all RGB frames from the ETRI-Activity3D dataset were resized from $1920 \times 1080$ to $456 \times 256$ pixels to reduce computational load. To emphasize person-centric action rather than background information, we employed an object detection model YOLOv11~\cite{YOLOv11} to localize the main person in each frame. Observing that detected bounding boxes were often smaller than $224$ pixels, we used a fixed cropping window of $224 \times 224$ pixels centered on the detected person, ensuring spatial consistency and preventing distortion.
Additionally, a uniform frame-sampling strategy was applied to standardize sequence lengths, with the optimal number of frames determined experimentally (detailed in Figure~\ref{fig:best_num_frames}). This approach significantly decreased memory and computational requirements, enabling efficient training with limited GPU resources.

After preprocessing the person-centric video clips, we adopted the standard cross-subject protocol for experimental evaluation on the ETRI-Activity3D dataset. Specifically, the entire subject pool was split into distinct training and validation sets: data from 67 subjects used for training and the remaining 33 subjects reserved for testing. The test set consisted of subject IDs \{3, 6, 9, 12, \ldots, 99\}, ensuring that all evaluations were conducted on previously unseen individuals. The data split was performed according to the official protocol provided by the ETRI-Activity3D dataset authors~\cite{FSA-CNN}.

\begin{figure}[h]
  \centering
  \includegraphics[width=0.85\linewidth]{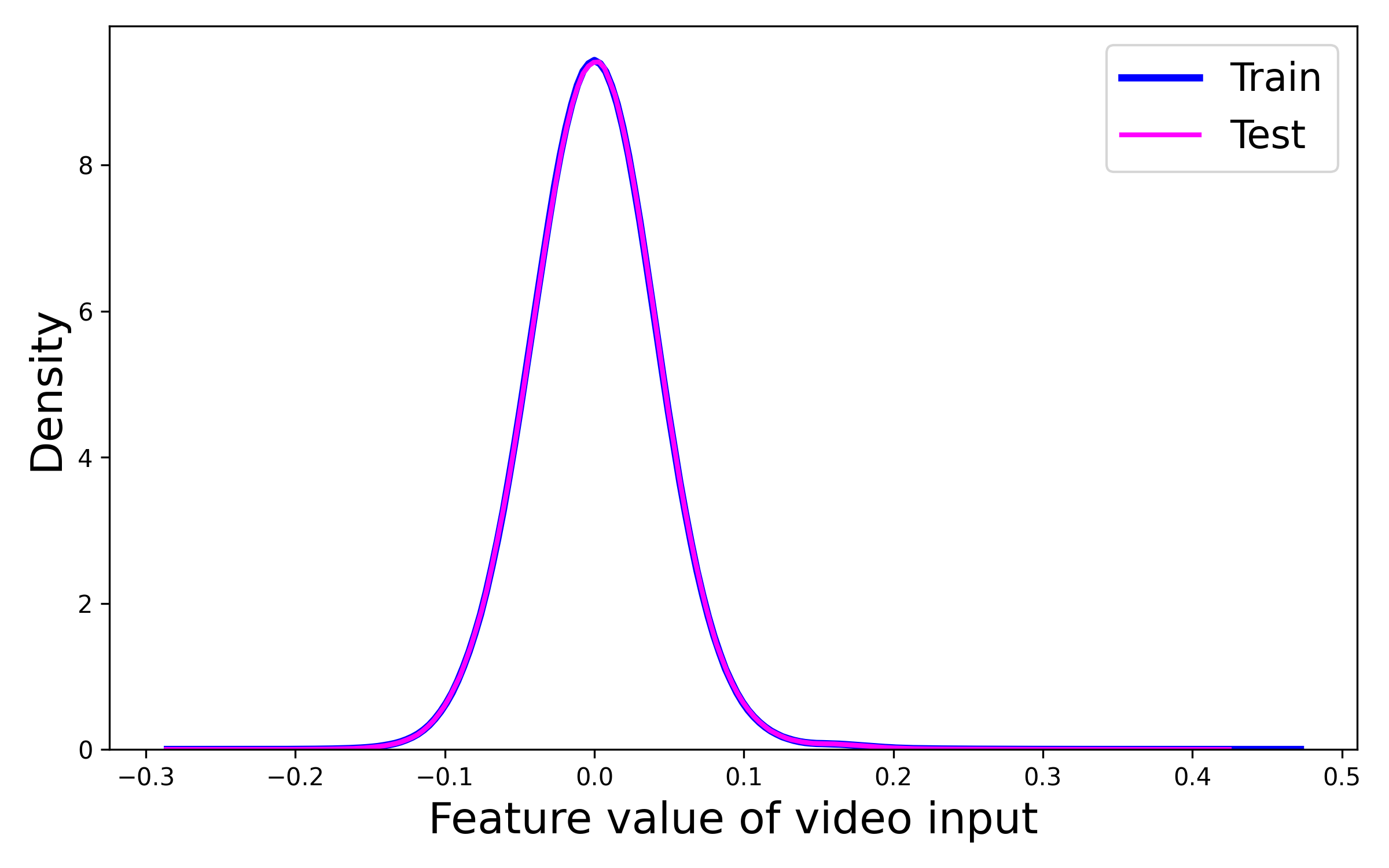}
  \caption{Distribution of video input values for train and validation sets}
  \label{fig:distribution}
\end{figure}

To further validate the reliability and consistency of our data partitioning, we visualized the empirical distribution of input features for both the training and test sets. As shown in Figure~\ref{fig:distribution}, the feature distributions of the training and test sets show complete overlap, confirming that the cross-subject split preserves the diversity and representativeness of the original dataset while ensuring no subject-level data leakage.

\section{Methodology}
\begin{figure}[h]
  \centering
  \includegraphics[width=\linewidth]{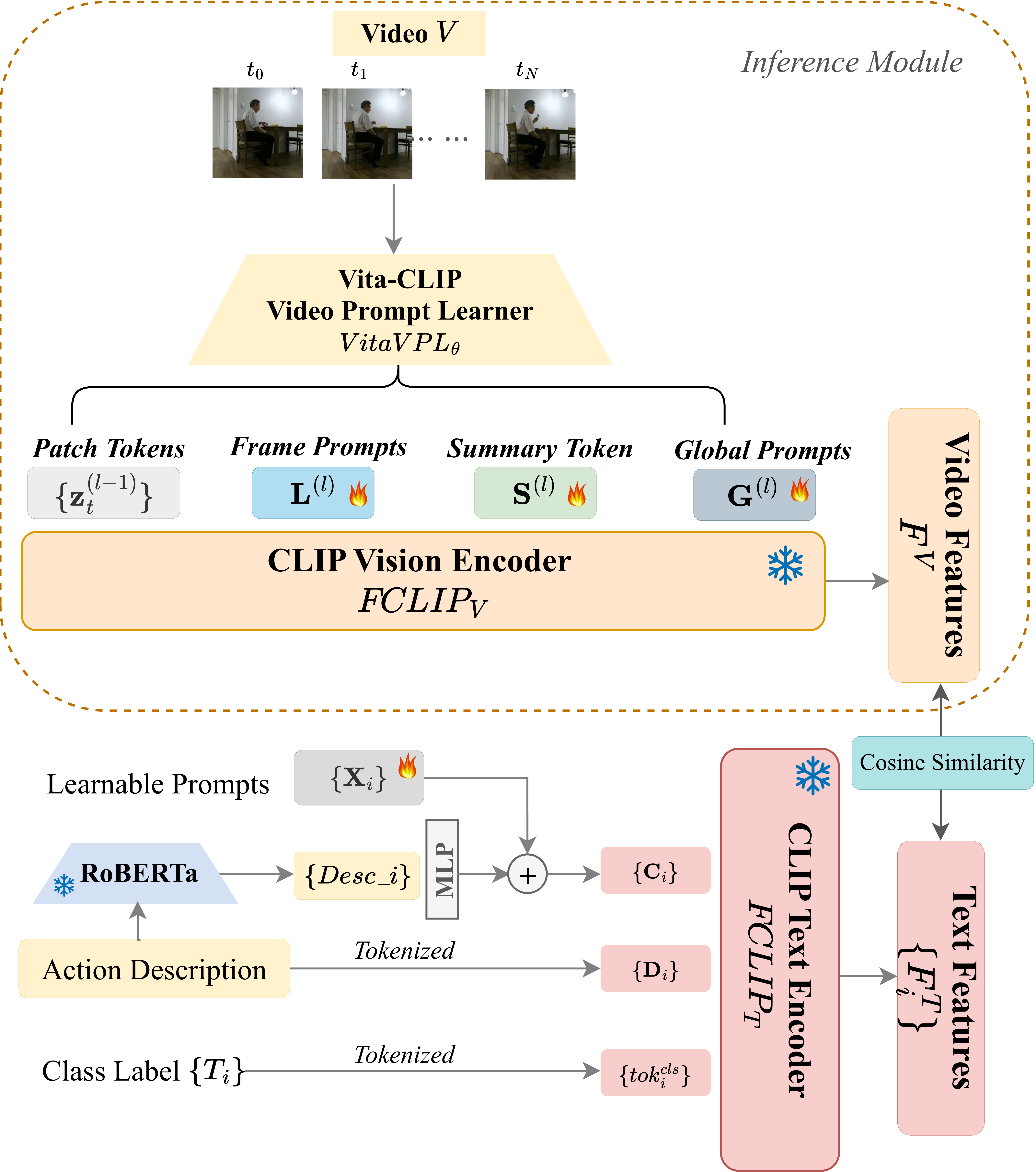}
  \caption{Architecture of the KRAST model. The top and bottom blocks depict the video and text encoding pipelines, respectively, which are jointly trained to align visual and textual features. During inference, only the video encoding component is evaluated to generate a representation, over which the final classification is performed.}
  \label{fig:architecture} 
\end{figure}

%\textcolor{red}{Recent studies, such as ActionCLIP~\cite{actionclip}, have shown that video action recognition can effectively leverage pretrained multimodal models like CLIP~\cite{radford2021learning} by introducing task-specific prompts with learnable parameters. While many works~\cite{CLIP4Clip,actionclip} perform end-to-end fine-tuning of the VLM using downstream video data, our approach instead focuses on prompt learning—adapting the pretrained model for general action recognition via knowledge-augmented prompts, without modifying the core encoder weights.}
Our approach leverages a large pre-trained vision–language model (VLM) and focuses on prompt learning—adapting the model for our action recognition task using knowledge-augmented prompts, without modifying the core encoder weights. Inspired by knowledge-aware prompt tuning (KAPT)~\cite{kapt}, we use action-specific textual descriptions as prior knowledge to condition the prompts, thereby improving human activity recognition (HAR) from monocular videos. This contrasts with prior works~\cite{CLIP4Clip,actionclip} that perform end-to-end fine-tuning of VLMs using downstream video data.
Fig. \ref{fig:architecture} illustrates the architecture of our model. During training, prompts for both text and video are jointly optimized to align their feature representations. A key aspect of our approach is the knowledge-aware initialization of text prompts, where we use the textual description of each action class. 
During inference, only video data are used: the video encoder generates a visual representation upon which the final classification is performed.
The remainder of this section is organized as follows. In Section~\ref{sect:C_D_learnable_prompts}, we introduce different forms of learnable prompts, namely continuous and discrete prompts.
Section~\ref{sect:SegKPT} presents our strategy for discrete prompting, while Section~\ref{sect:Visual_prompt} details the design of video prompts.
Finally, we describe  the classification process in Section~\ref{sect:action_classification}.

\subsection{Continuous and Discrete Text Prompts}
\label{sect:C_D_learnable_prompts}
%The semantic knowledge in the form of textual description on each action class or category are used to initialize text prompts.
Building on the concept of knowledge-aware prompt tuning (KAPT)~\cite{kapt}, we design two types of prompts: (i) \textit{continuous} prompts that capture broad contextual information about each class, and (ii) \textit{discrete} prompts derived from textual summaries of class descriptions. These descriptions are first generated using a large language model (ChatGPT) and then reviewed and improved manually to make sure they are clear and relevant. 
These text descriptions are built into learnable (continuous) or fixed (discreet) prompt vectors, which act as soft prompts that help the model focus on class-specific semantics when it is classifying.

%We further enhance this approach by conditioning prompts on action class names and relevant context. Specifically, 
For the continuous prompts, the final per-class prompt embedding $C_i$ is constructed as:
\[
C_i = \text{MLP}_T(\text{Desc}_i) + {X_i},
\] 
where $\text{Desc}_i$ is the semantic embedding of the $i$-th class description generated by  RoBERTa~\cite{liu2019roberta}, $\text{MLP}_T$ is a multi-layer perceptron, and $X_i$ is an additional learnable vector. This continuous prompt tuning (CPT) framework enables the model to capture both high-level semantic structure and class-specific variations.

Compared to baseline prompt-free models, we found that continuous prompts extracted from descriptive texts result in a more discriminative and coherent feature space in practice, making it easier to distinguish between action classes. However, KAPT~\cite{kapt} indicates that prompts trained on specific data may overfit to seen data. Building on this insight, we tokenize class descriptive texts into discrete prompts $\{D_i\}$ to better leverage semantic knowledge. Considering the 77-word context length limitation of the frozen CLIP ~\cite{radford2021learning} text encoder ($\text{FCLIP}_{\!T}$ in Fig.\ref{fig:architecture}), we craft two variants of discrete prompts to condense the text length.

    \paragraph{Keyword-wise Prompt Tuning (KeyPT):} For each action class, we extract a curated set of key attributes or descriptive phrases (e.g., ‘hand care activity’, ‘washing both hands with water and soap’, ‘rinsing under running water’) to serve as textual prompts. These keywords are \textbf{\underline{bold-faced and underlined}} within the textual descriptions in Table~\ref{tab:action_descriptions}. This approach condenses essential semantics into a compact form and promotes efficient knowledge transfer.

    \paragraph{Segmented Knowledge Prompt Tuning (SegKPT):} For action classes with longer or more descriptive annotations, we divide the full textual description into several meaningful segments, with each segment capturing a different aspect of the action. As illustrated in Table~\ref{tab:action_descriptions}, the first segment is generated using the hierarchical strategy (\textcolor{Red}{H}), the second segment is generated based on semantic attributes (\textcolor{OliveGreen}{S}), and the third segment is generated by the discriminative strategy (\textcolor{RoyalBlue}{D}). This segmentation yields the best empirical performance among the strategies we explored. In the following, we detail the design rationale and construction process for each strategy.

\subsection{{Strategies for SegKPT}}
\label{sect:SegKPT}
We developed three different strategies for the SegKPT. 
\begin{itemize}
    \item \textbf{Hierarchical strategy (H):} {To construct the first segment of the SegKPT prompts, we hierarchically group the 55 action classes based on their semantic similarity. The resulting multi-level categorization is shown in Table~\ref{tab:hierarchical_group}, where each action is assigned to a Level-1 category (e.g., food consumption, personal care, non-verbal communication), representing the broad semantic domain, and a Level-2 category (e.g., eating activity, face care activity, hand gesture), describing the fine-grained sub-group. We then use ChatGPT to generate representative descriptions for each action leveraging both the Level-1 cluster context and the Level-2 sub-cluster specificity. This ensures that the prompts capture global semantic coherence while highlighting distinctive details for each action.}

    \item \textbf{Semantic strategy (S):} In this strategy, we directly use ChatGPT to generate a concise semantic description for each action class based on its original annotation. These descriptions summarize the concept and key characteristics of each action, serving as interpretable prompts that retain clinically relevant context. The resulting sentences, shown in \textcolor{OliveGreen}{green} in Table~\ref{tab:action_descriptions}, provide a compact yet informative abstraction of each class to guide the vision-language model.

    \item \textbf{Discriminative strategy (D):} This strategy focuses on identifying key discriminative features to distinguish between action classes that are semantically similar or close in the embedding space. For example, actions such as “washing hands” and “washing a towel by hand” may appear visually similar but differ in subtle contextual cues. We use ChatGPT to generate prompts that highlight such distinctive attributes, such as the object being washed, the motion pattern, or the typical hand configuration, helping the model to separate closely related classes better. These discriminative cues are shown in \textcolor{RoyalBlue}{blue} in Table~\ref{tab:action_descriptions}.
\end{itemize}

\begin{figure}[h]
  \centering
  \includegraphics[width=\linewidth]{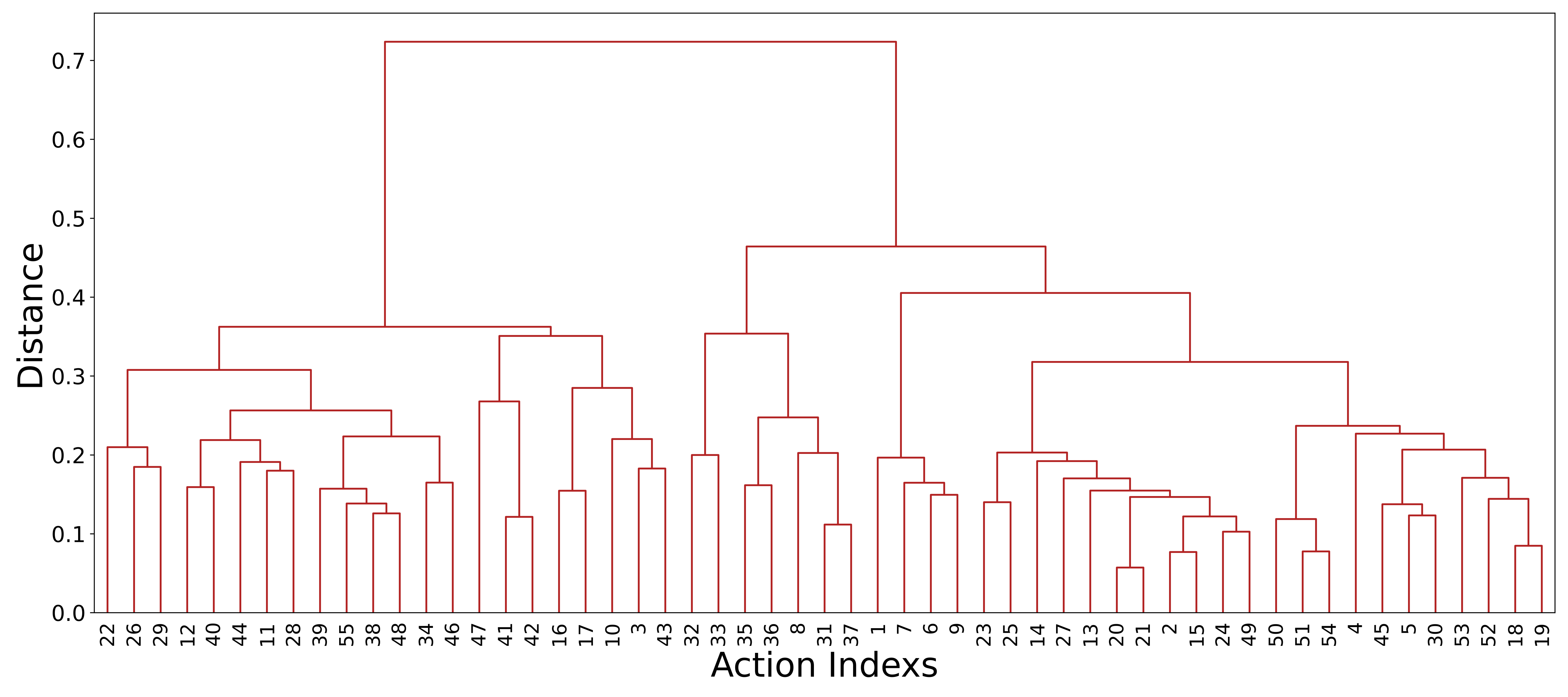}
  \caption{Representation of the relationship among the text embeddings of 55 action classes before the fine-tuning process}
  \label{fig:dendrogram_before_training}
\end{figure}

\begin{figure}[h]
  \centering
  \includegraphics[width=\linewidth]{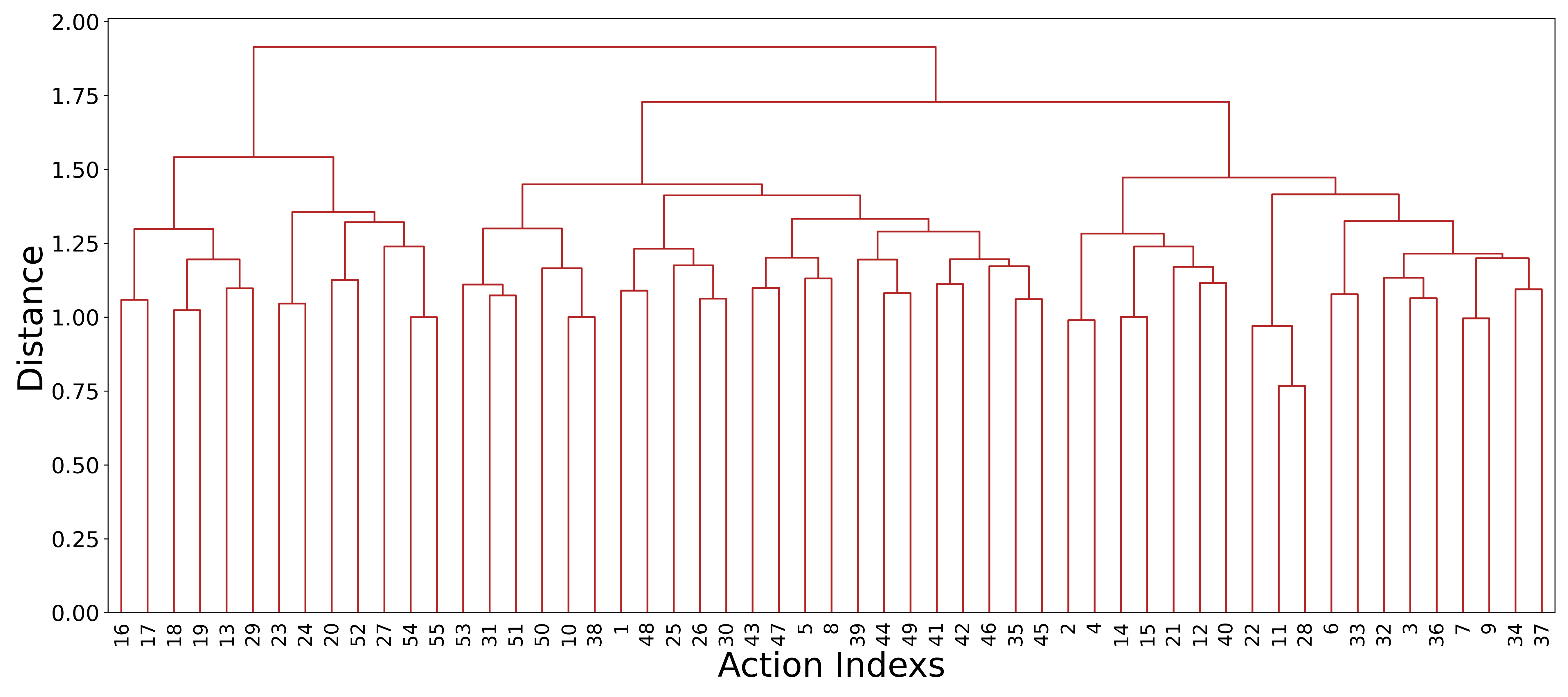}
  \caption{Representation of the relationship among the text embeddings of 55 action classes after the fine-tuning process}
  \label{fig:dendrogram_after_training}
\end{figure}

{To better understand the effect of fine-tuning, we visualize the similarity relationships among the text embeddings of all 55 action classes before and after the process, by using Ward's algorithm \cite{ward_algorithm}. It groups data points by minimizing the increase in total within-cluster variance, and represents the results as dendrograms, as shown in Fig.\ref{fig:dendrogram_before_training} and Fig.\ref{fig:dendrogram_after_training}. In this representation, actions that are grouped together early in the hierarchy are considered highly similar, whereas actions that merge only near the top of the tree exhibit weaker similarity. The height at which two clusters merge indicates the distance (or dissimilarity) between them.}

{We observe that the prompt-tuning process leads to a coherent clustering structure. Initially, many action classes are placed very close to each other (i.e. low height of the tree nodes), making it difficult to distinguish between semantically different actions. After tuning, the distances between unrelated actions become larger (e.g. 31:``using a remote control" and 37:``using a computer"), while semantically related actions (e.g., 11:``washing hands", 28:``washing a towel by hand", and 22:``washing dishes") are consistently grouped together. This indicates that fine-tuning improves the semantic separability of action classes, which in turn contributes to better recognition performance.}

\subsection{Prompting  Vision Encoder}
\label{sect:Visual_prompt}
On the video side, each frame of the input video $V$ goes through the tokenization of the Vision Transformer (ViT)~\cite{dosovitskiy2021an}, collectively forming a sequence of per-frame representations $z_t^{(0)}$. The visual prompts for the $l$-th layer of the frozen CLIP vision encoder $FCLIP_V$ are derived by applying a video prompt learner to the output of the previous layer $\{z_t^{(l-1)}\}$:
\[
[S^{(l)}, G^{(l)}, L^{(l)}]_{l=1,...,12} = \text{VitaVPL}_\theta(\{z_t^{(l-1)}\}),
\]
where $S^{(l)}$, $G^{(l)}$, and $L^{(l)}$ denote the learnable summary, global, and local prompt tokens at layer $l$, respectively, and $\text{VitaVPL}_\theta$ is the CLIP ViT encoder inserted with learnable prompts. The prompt tokens are appended to the per-frame representations and subsequently fed into $FCLIP_V$ to obtain the visual feature $F^V$:
\[
F^V = {FCLIP}_V([\{z_t^{(l-1)}\}, S^{(l)}, G^{(l)}, L^{(l)}]).
\]
\subsection{Action Classification}
\label{sect:action_classification}
We determine the class label of the visual feature $F^V$ by comparing it to the per-class text features $\{F^T_i\}$ encoded from the text encoder. The class with the highest similarity score is chosen as the label. The trainable components of our model are optimized using a contrastive loss to maximize the cosine similarity between class description-video pairs.

To address the class imbalance in our dataset (Fig.~\ref{fig:statistics}),  we apply a multi-class focal loss~\cite{focal-loss} that enhances the cosine similarity of positive pairs while down-weighting easy negatives.
The loss for text-video contrastive learning is formulated as:
\[
\mathcal{L}_k = \sum_{i=1}^{N_{\text{cls}}} \left[ -\alpha (1-p_i)^\gamma \right] y_i \log(p_i),
\]
where
\[
p_i = \frac{\exp(\langle F^T_i, F^V \rangle / \tau)}{\sum_{j=1}^{N_{\text{cls}}} \exp(\langle F^T_j, F^V \rangle / \tau)},
\]
and $y$ and $p_i$ denote the one-hot label and predicted probability of class $i$.
The cosine similarity between text and video feature pairs $\langle F^T_i, F^V \rangle$ is scaled by a learnable temperature parameter $\tau$, initialized to 0.01. We set the weighting factor $\alpha = 0.25$ and focusing parameter $\gamma = 2$ as in~\cite{focal-loss}.

After computing cosine similarities, we optionally apply temperature scaling using the learned $\tau$ and pass the scores through a softmax to obtain a probability distribution over the label set. 
During training, we keep the CLIP encoders frozen and update only the prompt parameters using a contrastive objective.
% NOT UNDERSTANDABLE: "test-time adaptation" below:
%During training, we update only the prompt parameters with our chosen objective (contrastive in our implementation), and we do not use test-time adaptation. 
During inference, the text embeddings are fixed and only the video input is encoded by the visual backbone. We then assign the video to the class with the highest probability.% and, if needed, also report a top-$K$ list. 

\section{Experiments and Results}
We conducted extensive experiments on the ETRI-Activity3D dataset to evaluate the performance of our model. 
%NOT UNDERSTANDABLE: Why zero-shot ?
%in a zero-shot action recognition setting. 

\subsection{Ablation Studies}
We analyzed the impact of two key factors on model performance: (i) the prompt-learning strategy and (ii) the number of video frames per sample.
These ablation experiments provide insight into how different knowledge injection strategies and temporal granularity affect recognition accuracy.

\paragraph{Prompt learning strategies}
We evaluated three prompt tuning approaches: Continuous Prompt Tuning (CPT), Keyword-based Prompt Tuning (KeyPT), and Segmented Knowledge Prompt Tuning (SegKPT). As shown in Table~\ref{tab:prompt_strategy}, the SegKPT strategy combining Hierarchical, Semantic, and Discriminative segments (S+H+D) achieved the best overall performance, with a top-1 accuracy of 95.22\%, F1-score of 0.946, and weighted F1-score of 0.952. Both KeyPT and CPT also improved upon the baseline, though with slightly lower performance gain compared to SegKPT. These results confirm that integrating diverse knowledge perspectives into prompts significantly enhances model understanding and classification capability.

\begin{table}
  \caption{Comparative analysis on model configurations. Model performance is evaluated using top-1 accuracy (\%), F1-score and weighted F1-score (“w.F1”). Best performances are highlighted in bold. \textcolor{Red}{(H): Hierarchical}, \textcolor{OliveGreen}{(S): Semantic}, \textcolor{RoyalBlue}{(D): Discriminative}}
  \label{tab:prompt_strategy}
  \begin{tabular}{lccc}
    \toprule
    Method & Accuracy & F1-score & weighted F1-score \\
    \midrule
    Baseline&75.31&0.734&0.753\\
    +CPT&87.14&0.862&0.871\\
    +KeyPT&92.46&0.918&0.924\\
    +SegKPT(textcolor{OliveGreen}{S}) & 87.38 & 0.865 & 0.873 \\
    +SegKPT(textcolor{OliveGreen}{S}+\textcolor{Red}{H}) & 93.70 & 0.924 & 0.937 \\
    +SegKPT(textcolor{OliveGreen}{S}+\textcolor{Red}{H}+\textcolor{RoyalBlue}{D}) &\textbf{95.22}&\textbf{0.946}&\textbf{0.952}\\
    \bottomrule
\end{tabular}
\end{table}

\begin{figure}[h]
  \centering
  \includegraphics[width=0.9\linewidth]{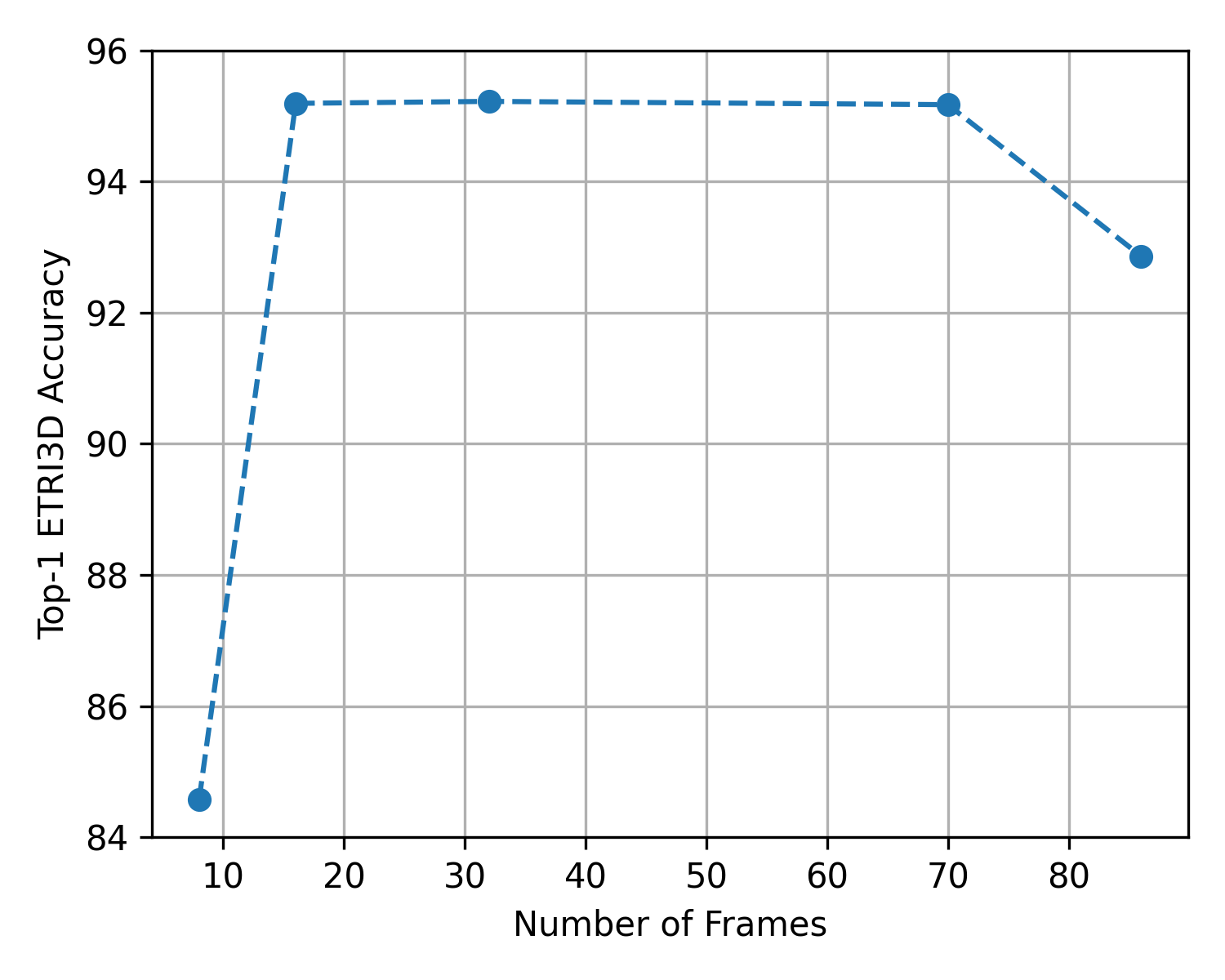}
  \caption{Top-1 classification accuracy on the ETRI-Activity3D dataset with different numbers of sampled video frames. The performance improves significantly up to 32 frames, then saturates or drops slightly with more frames.}
  \label{fig:best_num_frames}
\end{figure}

\paragraph{Effect of number of sampled frames}
To access the impact of temporal resolution, we tested the model with varying numbers of sampled frames per video: 8, 16, 32, 70, and 86. As shown in Figure~\ref{fig:best_num_frames}, accuracy improves significantly when increasing the number of frames from 8 to 16, reaching a peak of 95.22\% at 32 frames. However,
but additional frames yield only marginal gains or even degrade performance. 
This suggests that excessive frames may introduce redundant or redundant or uninformative data, 
%or less informative data
which can negatively affect the model’s attention mechanism while also increasing computational cost. We used 32-frame sampling in our work, as it offers an effective balance between temporal detail and computational efficiency.

\begin{figure}[h]
  \centering
  \includegraphics[width=\linewidth]{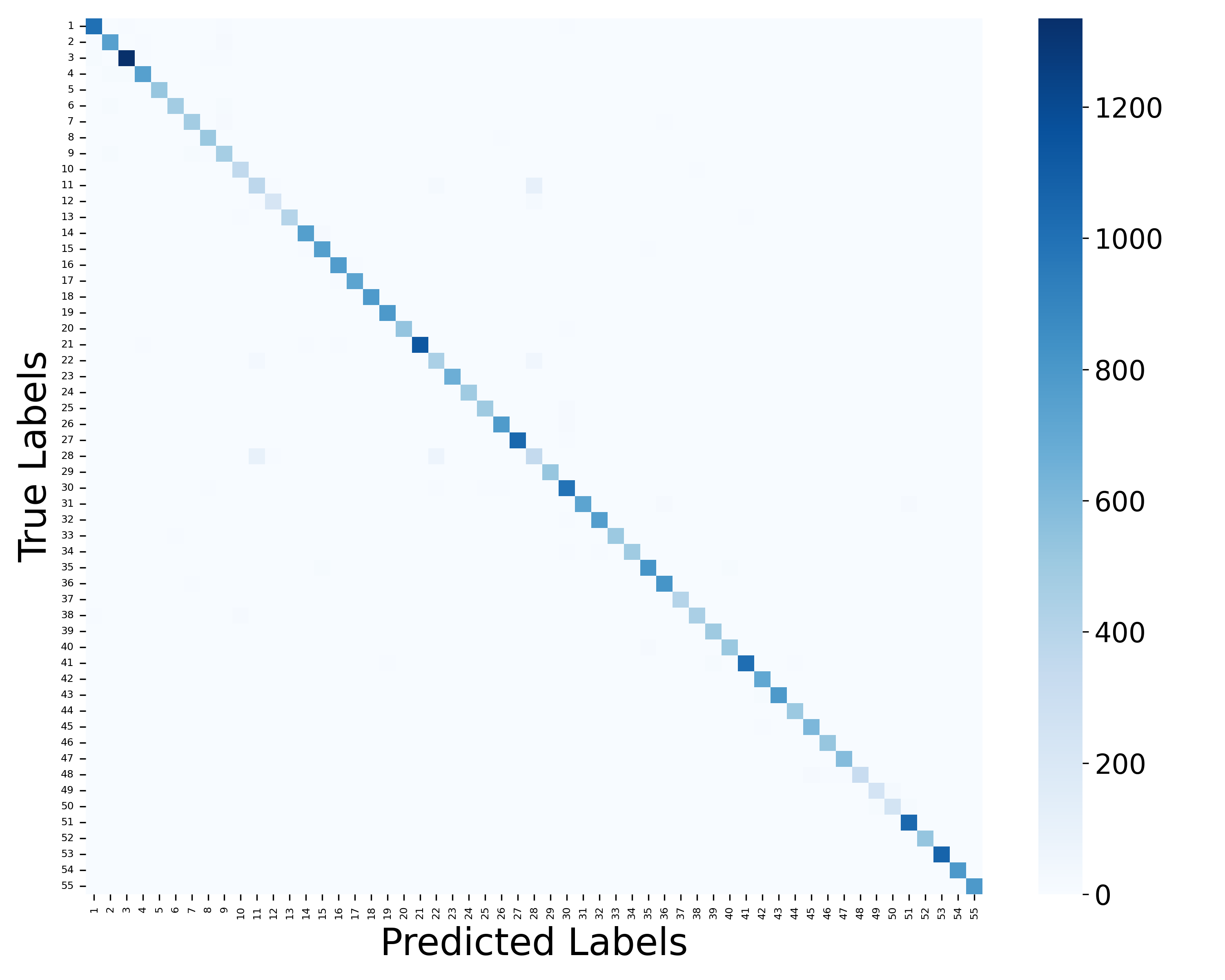}
  \caption{Confusion matrix of action classification results on the ETRI-Activity3D dataset.}
  \label{fig:confusion_matrix}
\end{figure}

Overall, the combination of SegKPT-- integrating hierarchical, semantic, and discriminative knowledge-- with an appropriate number of sampled frames leads to consistent performance gains. These results highlight the importance of both prompt design and temporal granularity in improving action recognition.% 'zero-shot' removed
Confusion matrix is provided in Figure~\ref{fig:confusion_matrix}.

\subsection{Comparison with state-of-the-art}

{To evaluate the effectiveness of the proposed model, we conducted a comparative analysis against several state-of-the-art (SOTA) methods previously tested on the ETRI-Activity3D~\cite{FSA-CNN} and NTU RGB+D 60~\cite{NTU60} datasets. Both datasets cover a wide range of indoor and daily activities, with ETRI-Activity3D specifically targeting elderly populations. By including both datasets in our evaluation, we provide a more comprehensive assessment and strengthen the reliability of the model’s performance. 
The selected SOTA baselines include Motif ST-GCN~\cite{MotifSTGCN}, HCN~\cite{HCN}, Evolution Pose Map\cite{EPM}, c-ConvNet\cite{ConvNet}, and FSA-CNN\cite{FSA-CNN}, with their reported results obtained from the experimental findings in~\cite{FSA-CNN}.}

{As presented in Table~\ref{tab:sota_comparison}, our model demonstrates superior performance compared to a wide range of existing approaches, including both single-modal and multi-modal baselines. Notably, whereas many of these methods rely on skeleton, depth, or multi-modal inputs—introducing additional pre-processing and computational burdens—our approach uses only RGB data while still attaining competitive accuracy. 
%Note that several prior methods depend on skeleton, depth, or combined modalities, which require complex preprocessing and higher computational cost. In contrast, our framework achieves competitive results using only RGB data.
For instance, on  ETRI-Activity3D dataset, the method delivers strong accuracy with a balanced precision/recall profile. On NTU RGB+D 60 dataset, the same prompt designs transfer effectively and remain competitive, indicating good generalization across different capture setups and subject populations. 
%Overall, structured textual knowledge improves class separability and robustness under limited supervision.

Together, these results highlight the effectiveness of integrating structured textual knowledge into prompt-based vision-language framework, which enables robust spatiotemporal representation learning without additional computational overhead.}

\begin{table}
  \centering
  \caption{Performance comparison of the proposed method and state-of-the-art approaches on the ETRI-Activity3D and NTU datasets. (Modalities: S=Skeleton, RGB=RGB video, D=Depth)}
  \label{tab:sota_comparison}
  \begin{tabular}{l@{\hspace{1em}}c@{\hspace{1em}}c@{\hspace{1em}}c}
    \toprule
    Method & Modalities & ETRI-3D & NTU RGB+D \\
    \midrule
    Motif ST-GCN~\cite{MotifSTGCN} & S & 89.9 & 84.2 \\
    HCN~\cite{HCN} & S & 88.0 & 86.5 \\
    Deep Bilinear Learning~\cite{DBL} & S & 88.4 & 85.4 \\
    Evolution Pose Map~\cite{EPM} & RGB+S & 93.6 & 91.7 \\
    c-ConvNet~\cite{ConvNet} & RGB+D & 91.3 & 82.6 \\
    FSA-CNN~\cite{FSA-CNN} & RGB & 90.1 & 87.2 \\
    FSA-CNN~\cite{FSA-CNN} & S & 90.6 & 88.1 \\
    FSA-CNN~\cite{FSA-CNN} & RGB+S & 93.7 & 91.5 \\
    \textbf{KRAST (ours)} & RGB & \textbf{95.22} & \textbf{90.2} \\
    \bottomrule
  \end{tabular}
\end{table}

\section{Conclusion and Future Work}
We presented KRAST, a novel knowledge-augmented prompting strategy for video-based action recognition with a pretrained vision–language model. By integrating structured domain knowledge into the prompting mechanism, our approach effectively bridges the visual and textual domains and yield significant performance improvements over image-only baselines and state-of-the-art methods. This work highlights the untapped potential of knowledge-augmented prompting for video understanding.

%Three different strategies for encoding the structural relationships of action classes into textual descriptions--hierarchical, semantic, and discriminative--are proposed and validated on two public datasets.
%The proposed continuous and structured discrete prompts consistently improve recognition performance over image-only baselines or state-of-the-art methods. 
%{This improvement demonstrates the strength of our approach in bridging vision and language for action recognition. By leveraging domain-specific knowledge through carefully designed prompts, the proposed model delivers reliable performance across different evaluation settings while maintaining computational efficiency and scalability compared to multi-modal alternatives.}

Our ultimate goal is the real-time deployment of our system for responsive human-robot interaction. 
This necessitates tackling the challenges of dynamic environments through robust methods for cross-dataset generalization, domain adaptation, and continual learning to ensure the model can adapt to novel actions.
%To ensure adaptability in dynamic environments, we will investigate cross-dataset generalization, domain adaptation, and continual learning, allowing the model to recognize novel actions over long-term use. %Finally, we consider incorporating additional modalities, such as depth or skeleton data when available, to improve robustness under challenging conditions. 
\begin{comment}
\begin{itemize}
  \item Deploy the system in real-time human–robot interaction for responsive, context-aware behaviors.
  \item Study cross-dataset generalization, domain adaptation, and continual learning so the model can adapt to novel actions over long-term use.
  \item Incorporate additional modalities such as depth or skeleton when available to improve robustness in challenging conditions.
  \item Optimize on-device efficiency for embedded and mobile robotic platforms.
\end{itemize}
\end{comment}

%%%%%%%%%%%%%%%%%%%%%%%%%%%%%%%%%%%%%%%%%%%%%%%%%%%%%%%%%%%%%%%%%%%%%%%%%%%%%%%%
\addtolength{\textheight}{-15cm}   % This command serves to balance the column lengths

%%%%%%%%%%%%%%%%%%%%%%%%%%%%%%%%%%%%%%%%%%%%%%%%%%%%%%%%%%%%%%%%%%%%%%%%%%%%%%%%


\onecolumn
\begin{table}[ht]
\centering
\caption{Summary of per-class textual descriptions used in prompt design. This table presents representative examples selected from the full set of 55 action classes in the dataset.
\textcolor{Red}{(H): Hierarchical} – Red sentences are generated using the hierarchical strategy, 
\textcolor{OliveGreen}{(S): Semantic} – Green sentences are generated using the semantic strategy, and 
\textcolor{RoyalBlue}{(D): Discriminative} – Blue sentences are generated using the discriminative strategy. 
The complete set of descriptions is available at \url{https://etri3ddescription.vercel.app/}.
}

\vspace{1.5em}

\renewcommand{\arraystretch}{2}
\begin{tabular}{|p{1cm}|p{3.2cm}|p{12cm}|}
\hline
\textbf{ID} & \textbf{Action} & \textbf{Description} \\ \hline
11 & Washing hands & \textcolor{Red}{(H) 
    A person is doing a \textbf{\underline{hand care activity}}, which is part of \textbf{\underline{personal care}}.
}

\textcolor{OliveGreen}{(S) 
    \textbf{\underline{Washing both hands with water and soap}}. The action involves rubbing palms together, scrubbing between fingers, cleaning the backs of hands and fingertips, and then \textbf{\underline{rinsing under running water}}. The movement is repetitive and focused around the sink area, often using both hands simultaneously in a coordinated pattern.
}

\textcolor{RoyalBlue}{(D) 
    \textbf{\underline{Both hands are actively scrubbed with soap}} under running water not just rinsing.
} \\ \hline

22 & Washing the dishes & \textcolor{Red}{(H) A person is doing a \textbf{\underline{dish cleaning activity}}, which is part of \textbf{\underline{cleaning}}.}

\textcolor{OliveGreen}{(S) \textbf{\underline{Washing the dishes}} while standing in front of a kitchen sink. One or both hands move in a \textbf{\underline{circular motion to scrub bowls or plates}} under running water. The person remains mostly in place, focusing on repetitive hand movements at the sink.}

\textcolor{RoyalBlue}{(D) \textbf{\underline{Hands or tools are used to scrub dishes}} under running water not just rinsing } \\ \hline

28 & Washing a towel by hands & \textcolor{Red}{(H) A person is doing a \textbf{\underline{laundry activity}}, which is part of \textbf{\underline{household chores}}.}

\textcolor{OliveGreen}{(S) \textbf{\underline{Cleaning a towel manually}} by \textbf{\underline{holding and rubbing the fabric with both hands}}, often while dipping it in water or soap. The action includes \textbf{\underline{twisting, squeezing, scrubbing, or wringing the towel}} to remove dirt. Movements are focused around the towel itself and usually happen near a basin or sink.}

\textcolor{RoyalBlue}{(D) \textbf{\underline{Hand washing a towel}} including twisting or scrubbing with both hands } \\ \hline

3 & Taking medicine & \textcolor{Red}{(H) A person is performing a \textbf{\underline{drinking activity}}, which is part of \textbf{\underline{food consumption}}.}

\textcolor{OliveGreen}{(S) Taking a \textbf{\underline{pill}} using fingers followed by cup drinking. Presence of a \textbf{\underline{pill bottle}}, hand motion toward mouth with small object, then immediate water intake to swallow.}

\textcolor{RoyalBlue}{(D) Distinct from simple drinking by the presence of a \textbf{\underline{pill and immediate water intake}} } \\ \hline

4 & Drinking water & \textcolor{Red}{(H) A person is performing a \textbf{\underline{drinking activity}}, which is part of \textbf{\underline{food consumption}}.}

\textcolor{OliveGreen}{(S) Only \textbf{\underline{drinking water from a cup}}. No sign of pill or pill bottle. The action includes \textbf{\underline{picking up the cup and tilting it to drink}}, without prior pill handling.}

\textcolor{RoyalBlue}{(D) No pill medication or food involved—only \textbf{\underline{water is consumed from the cup}}. } \\ \hline

2 & Pouring water into a cup & \textcolor{Red}{(H) A person is performing a \textbf{\underline{serving activity}}, which is part of \textbf{\underline{food consumption}}.}

\textcolor{OliveGreen}{(S) A person holds \textbf{\underline{a bottle or pitcher}} above a cup and carefully tilts it to \textbf{\underline{pour water}}. The liquid flows steadily into the cup without spilling. The cup stays on the table, and the person \textbf{\underline{does not bring it to their mouth}} or consume the liquid. No medicine, spoon, or other utensil is involved.}

\textcolor{RoyalBlue}{(D) \textbf{\underline{Water is poured}} without drinking cup remains on the table throughout the action. } \\ \hline

14 & Putting on cosmetics & \textcolor{Red}{(H) A person is doing a \textbf{\underline{cosmetic application activity}}, which is part of \textbf{\underline{personal care}}.}

\textcolor{OliveGreen}{(S) \textbf{\underline{Applying makeup to the face}} using hands, brushes, or sponges. The action includes touching areas like cheeks, forehead, eyes, or nose in repeated motions. The person may hold a \textbf{\underline{small mirror or makeup item}} and use \textbf{\underline{circular or tapping movements}} to apply products. Often done in front of a mirror while seated or standing.}

\textcolor{RoyalBlue}{(D) \textbf{\underline{Cosmetic products are applied}} using hands or tools focus on face not hair or body. } \\ \hline

15 & Putting on lipstick & \textcolor{Red}{(H) A person is doing a \textbf{\underline{cosmetic application activity}}, which is part of \textbf{\underline{personal care}}.}

\textcolor{OliveGreen}{(S) \textbf{\underline{Applying lipstick}} while seated in front of a mirror, holding the \textbf{\underline{lipstick with one hand}} and carefully \textbf{\underline{tracing the shape of both lips}}. The person looks into the mirror and moves the hand slowly across the lips in a precise motion.}

\textcolor{RoyalBlue}{(D) \textbf{\underline{Lipstick is applied specifically to lips}} not general makeup } \\ \hline

49 & Waving a hand & \textcolor{Red}{(H) A person is doing a \textbf{\underline{hand gesture}}, which is part of \textbf{\underline{non-verbal communication}}.}

\textcolor{OliveGreen}{(S) \textbf{\underline{Raising one arm}} and \textbf{\underline{moving the hand side to side}} in the air to greet or say goodbye. The motion is usually smooth, repeated a few times, and aimed toward another person. The arm is lifted to about head or shoulder level, and the hand swings clearly to attract attention.}

\textcolor{RoyalBlue}{(D) \textbf{\underline{One hand is raised and waved side to side}} as a greeting or farewell } \\ \hline

50 & Flapping a hand up and down (beckoning) & \textcolor{Red}{(H) A person is doing a \textbf{\underline{hand gesture}}, which is part of \textbf{\underline{non-verbal communication}}.}

\textcolor{OliveGreen}{(S) Raising one hand with the palm facing downward and \textbf{\underline{moving it up and down repeatedly}} to signal someone to come closer. The motion is vertical, short, and often done with fingers slightly bent or together. The arm stays in a fixed position while the hand moves, usually directed toward another person nearby.}

\textcolor{RoyalBlue}{(D) \textbf{\underline{Hand is flapped up and down to beckon}} not waving side to side. } \\ \hline

\multicolumn{1}{|c|}{
  $\begin{array}{c}
  \raisebox{0pt}{\scalebox{1.5}{$\cdot$}} \\[-0.2em]
  \raisebox{0pt}{\scalebox{1.5}{$\cdot$}} \\[-0.2em]
  \raisebox{0pt}{\scalebox{1.5}{$\cdot$}}
  \end{array}$
}
&
\multicolumn{1}{c|}{
  $\begin{array}{c}
  \raisebox{0pt}{\scalebox{1.5}{$\cdot$}} \\[-0.2em]
  \raisebox{0pt}{\scalebox{1.5}{$\cdot$}} \\[-0.2em]
  \raisebox{0pt}{\scalebox{1.5}{$\cdot$}}
  \end{array}$
}
& 
\multicolumn{1}{c|}{
  $\begin{array}{c}
  \raisebox{0pt}{\scalebox{1.5}{$\cdot$}} \\[-0.2em]
  \raisebox{0pt}{\scalebox{1.5}{$\cdot$}} \\[-0.2em]
  \raisebox{0pt}{\scalebox{1.5}{$\cdot$}}
  \end{array}$
} \\ \hline

\end{tabular}
\label{tab:action_descriptions}
\end{table}

\onecolumn
\begin{table}[ht]
\centering
\caption{ETRI-Activity3D: Action Classes Organized by Hierarchical Categories with Prompts }

\vspace{1.5em}

\renewcommand{\arraystretch}{2}
\begin{tabular}{|p{1cm} |p{3cm}|p{3cm}|p{3cm}|p{5.5cm}|}
\hline
\textbf{ID} & \textbf{Action} & \textbf{Level 1 Category} & \textbf{Level 2 Category} & \textbf{Prompt} \\
\hline

1 & eating food with fork & food consumption & eating activity & A person is performing an eating activity, which is part of food consumption.
 \\ \hline
2 & pouring water & food consumption & serving activity & A person is performing a serving activity, which is part of food consumption.
 \\ \hline
3 & taking medicine & food consumption & drinking activity & A person is performing a drinking activity, which is part of food consumption.
 \\ \hline
4 & drinking water & food consumption & drinking activity & A person is performing a drinking activity, which is part of food consumption.
 \\ \hline
5 & putting food in fridge & food-related activities & food storage & A person is engaged in food storage, which is part of food-related activities.
 \\ \hline
6 & trimming vegetables & food preparation & cutting activity & A person is performing a cutting activity, which is part of food preparation.
 \\ \hline
7 & peeling fruit & food preparation & peeling activity & A person is performing a peeling activity, which is part of food preparation.
 \\ \hline
8 & using gas stove & food preparation & cooking equipment use activity & A person is performing a cooking equipment use activity, which is part of food preparation.
 \\ \hline
9 & cutting vegetable & food preparation & cutting activity & A person is performing a cutting activity, which is part of food preparation.
 \\ \hline
10 & brushing teeth & personal care & dental care activity & A person is doing a dental care activity, which is part of personal care.
 \\ \hline
11 & washing hands & personal care & hand care activity & A person is doing a hand care activity, which is part of personal care.
 \\ \hline
12 & washing face & personal care & face care activity & A person is doing a face care activity, which is part of personal care.
 \\ \hline
13 & wiping face with towel & personal care & face care activity & A person is doing a face care activity, which is part of personal care.
 \\ \hline
14 & putting on cosmetics & personal care & cosmetic application activity & A person is doing a cosmetic application activity, which is part of personal care.
 \\ \hline
15 & putting on lipstick & personal care & cosmetic application activity & A person is doing a cosmetic application activity, which is part of personal care.
 \\ \hline
16 & brushing hair & grooming & hair care activity & A person is doing a hair care activity, which is part of grooming.
 \\ \hline
17 & blow drying hair & grooming & hair care activity & A person is doing a hair care activity, which is part of grooming.
 \\ \hline
18 & putting on a jacket & clothing and accessories & clothing management & A person is engaged in clothing management, which is part of clothing and accessories.
 \\ \hline
19 & taking off a jacket & clothing and accessories & clothing management & A person is engaged in clothing management, which is part of clothing and accessories.
 \\ \hline

\multicolumn{1}{|c|}{
  $\begin{array}{c}
  \raisebox{0pt}{\scalebox{1.5}{$\cdot$}} \\[-0.2em]
  \raisebox{0pt}{\scalebox{1.5}{$\cdot$}} \\[-0.2em]
  \raisebox{0pt}{\scalebox{1.5}{$\cdot$}}
  \end{array}$
}
& 
\multicolumn{1}{|c|}{
  $\begin{array}{c}
  \raisebox{0pt}{\scalebox{1.5}{$\cdot$}} \\[-0.2em]
  \raisebox{0pt}{\scalebox{1.5}{$\cdot$}} \\[-0.2em]
  \raisebox{0pt}{\scalebox{1.5}{$\cdot$}}
  \end{array}$
} 
& 
\multicolumn{1}{|c|}{
  $\begin{array}{c}
  \raisebox{0pt}{\scalebox{1.5}{$\cdot$}} \\[-0.2em]
  \raisebox{0pt}{\scalebox{1.5}{$\cdot$}} \\[-0.2em]
  \raisebox{0pt}{\scalebox{1.5}{$\cdot$}}
  \end{array}$
} 
&
\multicolumn{1}{|c|}{
  $\begin{array}{c}
  \raisebox{0pt}{\scalebox{1.5}{$\cdot$}} \\[-0.2em]
  \raisebox{0pt}{\scalebox{1.5}{$\cdot$}} \\[-0.2em]
  \raisebox{0pt}{\scalebox{1.5}{$\cdot$}}
  \end{array}$
} 
&
\multicolumn{1}{c|}{
  $\begin{array}{c}
  \raisebox{0pt}{\scalebox{1.5}{$\cdot$}} \\[-0.2em]
  \raisebox{0pt}{\scalebox{1.5}{$\cdot$}} \\[-0.2em]
  \raisebox{0pt}{\scalebox{1.5}{$\cdot$}}
  \end{array}$
} \\ \hline

55 & lying down	& body movements and postures & mobility activity & A person is performing a basic mobility activity, which is part of body movements and postures.
 \\ \hline

\end{tabular}
\label{tab:hierarchical_group}
\end{table}

\end{document}